\documentclass[]{template}

\usepackage[utf8]{inputenc}             
\usepackage[T1]{fontenc}                
\usepackage{url}                        
\usepackage{booktabs}                   
\usepackage{multirow}
\usepackage{colortbl}
\usepackage{multicol}
\usepackage{amsfonts}                   
\usepackage{nicefrac}                   
\usepackage{microtype}                  
\usepackage[dvipsnames]{xcolor}         

\usepackage{latexsym}

\usepackage{graphicx}
\usepackage{float}
\usepackage{subcaption}
\usepackage{wrapfig}
\usepackage{lipsum}
\usepackage{hyphenat}

\usepackage{bm}

\usepackage{tabularx} 
\usepackage{ragged2e} 
\newcolumntype{L}{>{\RaggedRight\hangafter=1\hangindent=0em}X}

\usepackage{enumitem}

\usepackage{amsmath}
\usepackage{amssymb}
\usepackage{mathtools}
\usepackage{amsthm}

\usepackage{listings}
\usepackage{placeins}
\usepackage{array}
\usepackage{cuted}
\usepackage{capt-of}
\usepackage{paracol}

\setboolean{logo}{true}    

\usepackage[linesnumbered,ruled,vlined]{algorithm2e}

\hypersetup{
    colorlinks=true,
    linkcolor=red,
    citecolor=Cerulean,
    filecolor=magenta,      
    urlcolor=magenta,
}

\usepackage[capitalize,noabbrev]{cleveref}
\crefname{section}{§}{§§}
\Crefname{section}{§}{§§}

\usepackage{calligra}
\DeclareMathAlphabet{\mathcalligra}{T1}{calligra}{m}{n}

\usepackage{pifont}

\theoremstyle{plain}

\theoremstyle{definition}

\theoremstyle{remark}

\renewcommand{\paragraph}[1]{\vspace{1mm}\noindent\textbf{#1}}

\DeclareCaptionLabelFormat{cont}{#1~#2\alph{ContinuedFloat}}
\usepackage[most]{tcolorbox}
\tcbset{
  promptbox/.style={
    top=10pt,
    colback=lightgray!20,
    colframe=Black,
    colbacktitle=NavyBlue,
    enhanced,
    center,
    attach boxed title to top center={yshift=-0.1in,xshift=0.0in},
    boxed title style={boxrule=0pt,colframe=white,},
  }
}
\newtcolorbox{promptbox}[2][]{promptbox, title=#2,#1}
\tcbset{
  takeawaybox/.style={
    top=10pt,
    colback=lightgray!20,
    colframe=Black,
    colbacktitle=BurntOrange,
    enhanced,
    center,
    attach boxed title to top center={yshift=-0.1in,xshift=0.0in},
    boxed title style={boxrule=0pt,colframe=white,},
  }
}
\newtcolorbox{takeawaybox}[2][]{takeawaybox, title=#2,#1}
\tcbset{
  observationbox/.style={
    top=10pt,
    colback=lightgray!20,
    colframe=Black,
    colbacktitle=YellowGreen,
    enhanced,
    center,
    attach boxed title to top center={yshift=-0.1in,xshift=0.0in},
    boxed title style={boxrule=0pt,colframe=white,},
  }
}
\newtcolorbox{observationbox}[2][]{observationbox, title=#2,#1}

\usepackage{xspace}

\newcommand\blfootnote[1]{%
  \begingroup
  \renewcommand\thefootnote{}\footnote{#1}%
  \addtocounter{footnote}{-1}%
  \endgroup
}

\usepackage{CJK}

\title{AgentCompass: A Unified Evaluation Infrastructure for Agent Capabilities}

\author{AgentCompass Team, Shanghai AI Laboratory$^{*}$}

\begin{abstract}

As Large Language Models (LLMs) evolve into autonomous agents, the need for unified evaluation infrastructure becomes critical. 
However, current evaluation pipelines remain highly fragmented and tightly coupled, hindering reproducibility and causing redundant engineering.
To address this, we introduce AgentCompass, an open-source, lightweight, and extensible infrastructure for evaluating LLM-based agents. 
AgentCompass organizes the evaluation process around three independent components, namely \textit{Benchmark}, \textit{Harness}, and \textit{Environment}, thereby enabling flexible configurations without requiring the reimplementation of complex execution logic. 
Furthermore, it features a fault-tolerant asynchronous runtime and comprehensive trajectory analysis tools to transparently diagnose nuanced failure modes like reward-hacking. 
Natively supporting over 20 benchmarks across five capability dimensions, AgentCompass provides the community with a scalable and reproducible infrastructure for advancing agent research.

\end{abstract}

\begin{document}

\blfootnote{$*$ Corresponding: \{zhudongsheng, mazerun, zhangqi\}@pjlab.org.cn \\
\makebox[2.4em][l]{} Code is at \url{https://github.com/open-compass/AgentCompass}
}

\maketitle

\section{Introduction}

\begin{wrapfigure}{r}{0.5\textwidth}
  \centering
  \includegraphics[width=\linewidth]{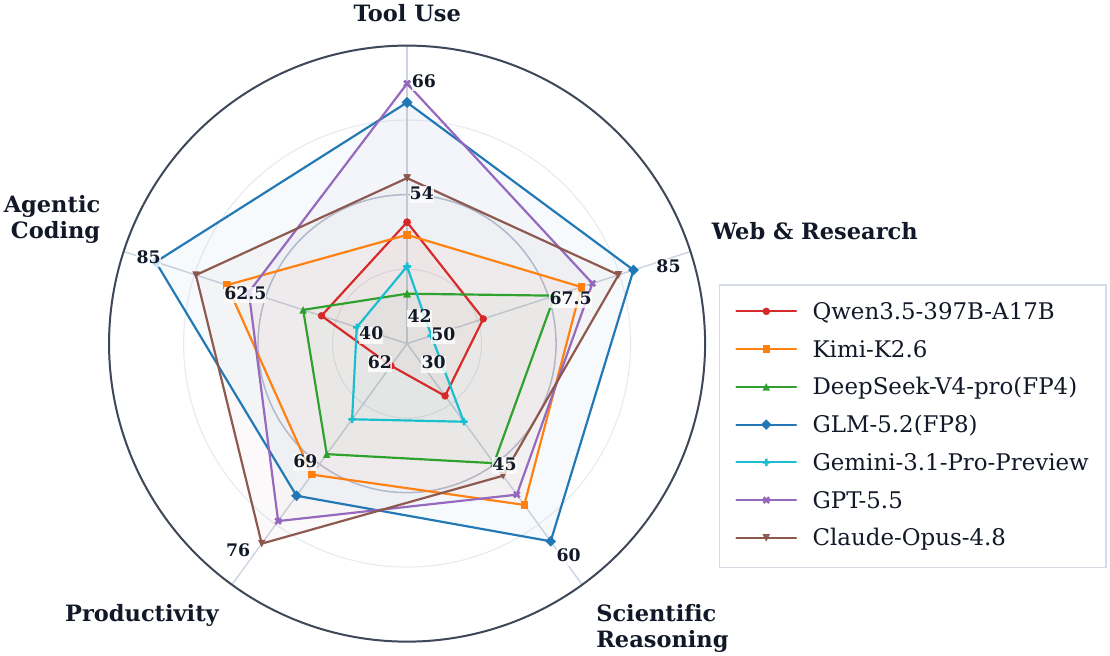}
  \caption{Capability profiles of representative models across the five core evaluation dimensions.}
  \label{fig:capability-radar}
\end{wrapfigure}

The paradigm of Large Language Models (LLMs) is rapidly shifting from instruction-following text generators to the foundation of LLM-based agents capable of complex reasoning, planning, and tool interaction in dynamic environments~\citep{wu2024autogen, li2023camel}. 
As LLM-based agents continue to advance in capability, corresponding evaluation methodologies must be developed in tandem to ensure rigorous assessment of these increasingly multi-faceted systems.

However, the current agent evaluation landscape is highly fragmented, suffering from a severe infrastructural deficit. 
While numerous specialized benchmarks have emerged to assess specific capabilities—such as tool invocation \cite{yao2024tau} and deep research \cite{mialon2023gaia}—they operate as isolated evaluation suites. 
This forces researchers to repeatedly configure heterogeneous execution environments, data formats, and scoring protocols. 
Such redundant engineering not only hinders efficiency but also compromises reproducibility due to inconsistent baseline implementations. 
Furthermore, existing general-purpose infrastructure either lack native support for interactive agent workflows \cite{2023opencompass, duan2024vlmevalkit} or restrict their scope to narrow domains like coding \cite{Harbor_Framework}. 
The community urgently needs a unified, extensible infrastructure to streamline agent evaluation.

To bridge this gap, we introduce \textbf{AgentCompass}, an open-source, lightweight, and extensible evaluation infrastructure designed to support systematic evaluation of LLM-based agents.
Its core design philosophy lies in decoupling three typically entangled components: \textit{Benchmark}, \textit{Harness}, and \textit{Environment}.
By abstracting the harness into an independent layer, AgentCompass transforms rigid evaluation pipelines into flexible $\text{benchmark} \times \text{harness} \times \text{Environment}$ configurations. 
This architecture allows researchers to evaluate different agents on a shared benchmark, compare alternative interaction protocols, or integrate new datasets without reimplementing complex execution logic. 
As illustrated in Figure~\ref{fig:capability-radar}, this unified approach readily enables the comprehensive profiling of representative models across multiple core capability dimensions.

Built upon this modular abstraction, AgentCompass standardizes critical operations such as task dispatch, environment interaction, and metric aggregation. 
It features an asynchronous runtime optimized for parallelizing long-running agent trajectories. 
Beyond execution, the infrastructure provides rich trajectory-level analysis and visualization, capturing intermediate actions, tool calls, and environment feedback to help developers transparently diagnose failure modes. 
Having already served as the foundational evaluation infrastructure for the Intern-S agent series \cite{zou2026interns1proscientificmultimodalfoundation}, AgentCompass offers a scalable, reproducible, and highly accessible toolkit for advancing agent research.
\section{Related Work}

As agent capabilities continue to expand, evaluation must move beyond LLM benchmarks, giving rise to a broad yet fragmented landscape of agent-specific benchmarks and infrastructure.

\paragraph{Benchmarks.} Comprehensive benchmarks assess holistic performance via diverse interactive environments~\citep{liuagentbench}, challenging real world questions~\citep{mialon2023gaia,wei2025browsecomp,zhou2025browsecomp,phan2025humanity,zhai2026hle,gupta2026deepsearchqa}, and game-playing scenarios~\citep{iclr/WuTML24}, with specialized variants for multi-agent collaboration~\citep{zhu2025multiagentbench, hyun2025crew}. 
Other benchmarks provide granular analysis of tool and function calling~\citep{patil2025bfcl, yao2024tau, barres2025tau, shi2026tau}, scientific reasoning and research coding~\citep{wang2026frontierscience,xu2025probing,xu2026researchclawbench,tian2024scicode}, productivity-oriented skill execution~\citep{li2026skillsbench,pinchbench2026skill,patwardhan2025gdpval}, and agentic capabilities in software engineering~\citep{jimenez2024swebench, deng2025swe, yang2025swesmith,huang2026deepswe}, live coding~\citep{CodedotAI2023aider}, and terminal operations~\citep{tbench_2025, merrill2026terminal}.

\paragraph{Infrastructure.} To manage these diverse evaluations, supporting infrastructure has emerged. 
Some provide analytical dashboards~\citep{nips/MaZZYYJLKH24}, unit-testing frameworks~\citep{ConfidentAI2023deepeval}, or commercial tracing tools~\citep{LangChainLangSmith}. 
More recently, unified frameworks have been proposed to streamline development and evaluation. 
AgentGym~\citep{xi2024agentgym} provides infrastructure for evolving agents across diverse environments. 
Harbor~\citep{Harbor_Framework} further targets agent-oriented evaluation, with particular emphasis on coding and skill-based scenarios. 
For multi-agent systems, MASLab offers a unified and comprehensive codebase~\citep{ye2025maslab}.
Beyond agent evaluation, general-purpose frameworks also support broader model assessment: EvalScope~\citep{evalscope_2024} covers multiple evaluation modes and backends; OpenCompass~\citep{2023opencompass} targets broad reasoning and safety evaluation; and VLMEvalKit~\citep{duan2024vlmevalkit} focuses on image-text multimodal evaluation.

\section{Framework}
\label{sec:framework}

\begin{figure*}[t]
  \centering
  \includegraphics[width=\textwidth]{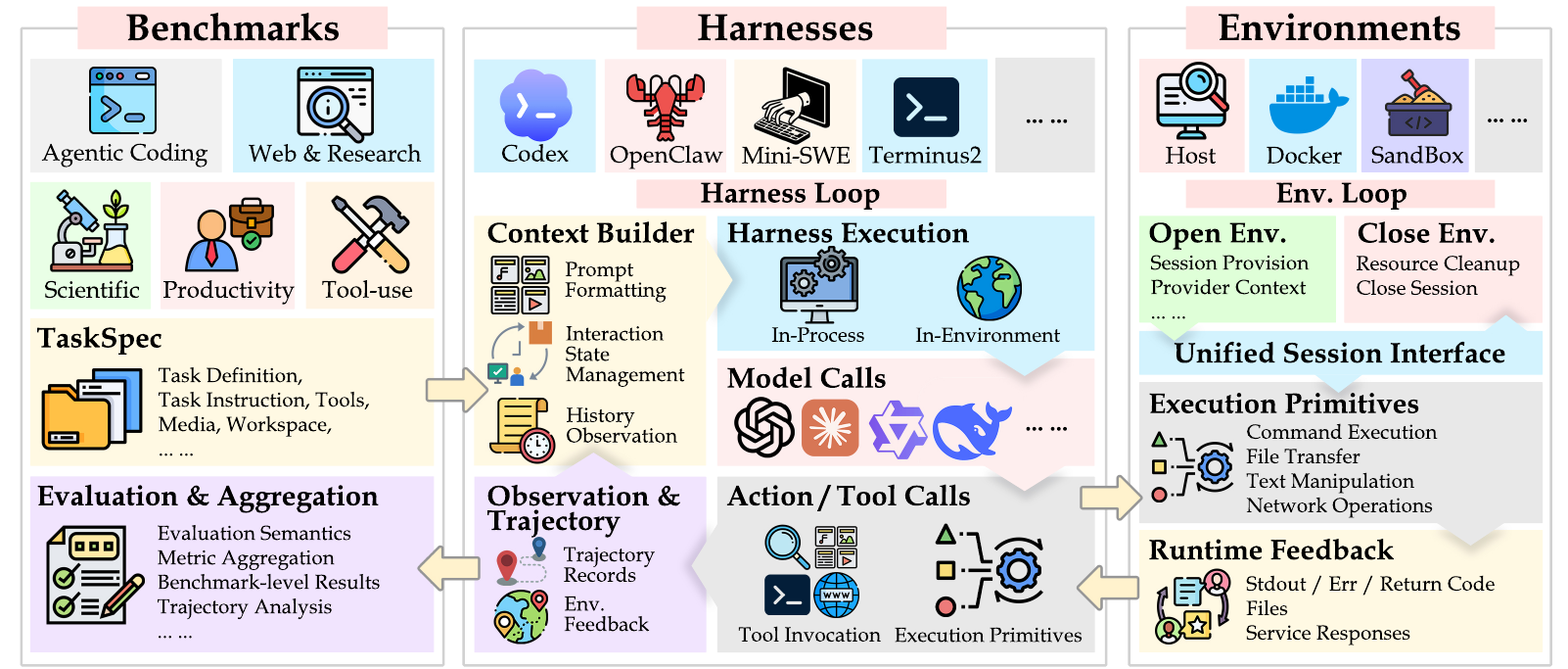}
  \caption{Overview of the AgentCompass architecture. The framework fully decouples Benchmarks, Harnesses, and Environments to enable flexible and composable agent evaluation.}
  \label{fig:overview}
\end{figure*}

AgentCompass modularizes benchmark-specific evaluation scripts into composable components connected by stable protocols. 
As shown in Figure~\ref{fig:overview}, each evaluation consists of a \emph{benchmark}, \emph{harness}, and \emph{environment}, with optional recipes and analyzers. 

\subsection{Design Overview}
\label{sec:design_overview}

In AgentCompass, an evaluation run is specified through a declarative \texttt{RunRequest}. 
The request separates the substantive objects of evaluation from the operational choices used to execute it. 
Specifically, \texttt{BenchmarkSpec} defines the task and evaluation metrics, \texttt{HarnessSpec} defines the agent procedure used to interact with each task, \texttt{EnvironmentSpec} identifies the execution context, and \texttt{ModelSpec} describes the model endpoint, credentials, inference parameters, and supported API protocols. 
Runtime options that govern how the evaluation is executed are placed in a separate \texttt{ExecutionSpec}, without changing the semantic definition of the evaluation itself.

This separation clarifies the configurable parts of an evaluation. 
For example, one may compare several harnesses on the same benchmark, reuse a harness across different benchmarks, or evaluate the same model through different interaction protocols without rewriting benchmark code. 
AgentCompass resolves these choices through lightweight, decorator-based registries for benchmarks, harnesses, environments, recipes, and analyzers. 
As a result, new components can be introduced by registering their implementations locally, rather than by modifying the central runtime or adapting unrelated modules. 
The resulting abstraction replaces fixed benchmark-specific pipelines with composable $\text{benchmark} \times \text{harness} \times \text{environment}$ configurations, while preserving a clear boundary between evaluation semantics and execution mechanics.

\subsection{Core Components and Protocols}
\label{sec:core_components}

AgentCompass improves modularity and extensibility by decomposing evaluation into protocol-driven components, replacing tightly coupled, benchmark-specific scripts with flexible architecture.

\paragraph{Benchmark.} 
The benchmark component encapsulates dataset-specific logic. 
It loads raw data into a uniform \texttt{TaskSpec}, prepares task materials, and computes final scores via an \texttt{evaluate} function. 
The grading mechanism supports deterministic matching, execution-based verification, and LLM-as-judge scoring. 
To separate agent rollout from evaluation, the benchmark defines a scorer execution mode: \texttt{none} for in-memory verification, \texttt{reuse} for the agent's existing workspace, or \texttt{fresh} for a clean, isolated test environment (e.g., applying an agent-generated patch to a pristine repository).

\paragraph{Harness.} 
The harness serves as the operational wrapper that instantiates a LLM into an interactive agent. 
It orchestrates the agent's internal logic, including prompt formatting, interaction state management, multi-turn tool invocation, and provider-specific API handling. 
Acting as a mediator, the harness consumes standardized task materials and coordinates with the model API and environment, entirely shielding the benchmark from agent-specific implementation details. 
The framework supports both in-process harnesses, where the interaction loop is managed natively by AgentCompass, and in-environment harnesses, which execute external agent frameworks as subprocesses within a sandbox. 
This accommodates a wide spectrum of systems, ranging from simple single-turn inference scripts to complex autonomous coding agents.

\paragraph{Environment.} 
The environment provides the isolated execution context and system primitives necessary for agent interaction and code verification. 
It exposes a unified session interface for executing commands, transferring files, manipulating text, and provisioning network services. 
Crucially, it serves as the security and isolation boundary for external agent execution. 
By abstracting the backend infrastructure, AgentCompass allows the same benchmark and harness configurations to run consistently across local host processes, local Docker containers, or distributed cluster instances. 
This allows researchers to scale evaluations from local development to large clusters without modifying the underlying code.

\paragraph{Protocol Abstractions} 
Component interoperability is guaranteed through two levels of strict protocol abstraction. 
First, models are represented purely as declarative API specifications (defining endpoints, parameters, and supported formats) rather than being tightly coupled to a monolithic client gateway. 
Second, data exchange between benchmarks and harnesses is governed by a standardized material protocol. 
Benchmarks compile each \texttt{TaskSpec} into a \texttt{PreparedTask} (bundling prompts, tools, media, and expected outputs), while the harness processes this structure to yield a uniform \texttt{RunResult} containing the final prediction, score, and the complete recorded trajectory. 
This explicit data contract eliminates the need for cross-modification when introducing new benchmarks or agents into AgentCompass.

\subsection{Asynchronous Execution Runtime}
\label{sec:execution_runtime}

The AgentCompass runtime is specifically designed to handle the high I/O overhead typical of agent-environment interactions. 
Users initiate an evaluation via the CLI or Python SDK by constructing a \texttt{RunRequest}. 
The runtime then dynamically resolves the registered components, prepares the task materials, and distributes tasks through an asynchronous dispatcher built on \texttt{asyncio}. 
By configuring concurrency limits, the infrastructure efficiently manages multiple parallel, long-running agent trajectories without blocking.
Furthermore, the runtime natively supports fault tolerance and state persistence. 
Partial results and structured progress events are saved incrementally. 
If an evaluation is interrupted, the infrastructure can resume seamlessly: it skips completed tasks and re-executes only the samples that encountered retryable failures. 
This incremental execution mechanism is crucial for cost-effective evaluation of autonomous agents.

\subsection{Trajectory Tracking and Behavioral Analysis}
\label{sec:trajectory_analysis}

While traditional benchmarks often collapse agent performance into a single scalar metric, AgentCompass records a comprehensive, versioned trajectory for each task. 
A trajectory captures the complete interaction sequence, including the agent's reasoning process, issued tool calls, environment feedback, and granular metrics (e.g., token consumption, inference latency, and stop reasons). 
This uniform schema standardizes interaction histories across heterogeneous models and benchmarks.

To facilitate in-depth diagnostics, AgentCompass provides a pluggable analyzer layer. 
Analyzers automatically process trajectories to extract structured insights, flagging problematic samples and categorizing errors into model-side, environment-side, or framework-side failures. 
Built-in analyzers can systematically detect anomalies such as output truncation, latency spikes, and repetitive generation loops. 
These trajectory-derived statistics help researchers to transparently pinpoint specific failure modes rather than solely observing overall performance degradation.

\begin{table*}[t]
\centering
\small
\caption{
Summary of the over 20 built-in benchmarks supported by AgentCompass,
categorized across key capability dimensions.
}
\label{tab:benchmarks_appendix}

\begin{tabularx}{\textwidth}{@{}>{\raggedright\arraybackslash}p{0.24\textwidth}>{\raggedright\arraybackslash}X@{}}
\toprule
\textbf{Capability Dimension} & \textbf{Benchmarks} \\
\midrule
Tool Use & Tau-bench~\citep{yao2024tau}, Tau2-bench~\citep{barres2025tau}, Tau3-bench~\citep{shi2026tau} \\
\addlinespace
Web \& Research
& BrowseComp~\citep{wei2025browsecomp}, BrowseComp-ZH~\citep{zhou2025browsecomp}, DeepSearchQA~\citep{gupta2026deepsearchqa}, GAIA~\citep{mialon2023gaia}, HLE~\citep{phan2025humanity}, HLE-Verified~\citep{zhai2026hle} \\
\addlinespace
Scientific Reasoning
& FrontierScience~\cite{wang2026frontierscience}, SciCode~\cite{tian2024scicode}, SGI-Bench(Deep Research)~\cite{xu2025probing}, ResearchClawBench~\cite{xu2026researchclawbench} \\
\addlinespace
Agentic Coding
& SWE-bench-Verified~\cite{yang2024sweagent}, SWE-bench-Pro~\cite{deng2025swe}, SWE-bench-Multilingual~\cite{yang2025swesmith}, Terminal-bench@2, Terminal-bench@2-Verified, Terminal-bench@2.1~\cite{tbench_2025} \\
\addlinespace
Productivity
& GDPVal-AC~\cite{patwardhan2025gdpval}, SkillsBench~\cite{li2026skillsbench}, PinchBench~\cite{pinchbench2026skill} \\
\bottomrule
\end{tabularx}

\end{table*}

\subsection{Extensibility and Reproducibility}
\label{sec:extensibility}

\paragraph{Extensibility.} 
AgentCompass employs a lightweight, registry-based architecture to minimize coupling. 
Introducing a new benchmark only requires a protocol-compliant subclass and local registration, with optional scoring functions and setup recipes. 
Because components are decoupled, existing harnesses and environments remain untouched. 
Model API protocols are likewise handled where they are consumed rather than through a monolithic client. 
This design lowers the barrier to integrating new capabilities within a unified codebase.

\paragraph{Reproducibility.} 
AgentCompass ensures strict tracking of evaluation provenance. 
Each run is persisted by benchmark and model, storing exact configurations, task-level logs, and aggregated summaries. 
AgentCompass distinguishes semantic parameters that alter agent behavior from execution parameters such as concurrency, so execution-only changes do not invalidate results. 
By isolating retryable failures from completed tasks, evaluations remain auditable and restartable at scale.

\subsection{Supported Benchmarks and Harnesses}
\label{sec:supported}

As of July 2026, AgentCompass natively integrates over 20 benchmarks spanning five core capability dimensions: tool use, web \& research, scientific reasoning, agentic coding, and productivity (summarized in Table~\ref{tab:benchmarks_appendix}). 
Rather than being limited to static question-answering, the integrated benchmarks encompass highly interactive and long-horizon environments, such as repository-level code modifications (e.g., SWE-bench variants) and complex open-ended web navigation (e.g., GAIA, HLE). 
AgentCompass also supports specialized evaluation paradigms; among them, GDPVal-AC is a custom AgentCompass variant that employs an agentic judger (via OpenClaw) to perform pair-wise evaluation against a Claude-Opus-4.8 baseline.

\begin{wraptable}{r}{0.5\textwidth}
\centering
\small
\begin{tabular}{ll}
\toprule
\textbf{Harness} & \textbf{Driven Agent / Workflow Type} \\
\midrule
OpenAI Chat Wrapper & Direct chat-completion \\
Naive Search Agent & Native deep-search loop \\
Claude Code & Claude Code CLI \\
Codex & OpenAI Codex CLI \\
OpenHands & OpenHands agent \\
OpenClaw & OpenClaw CLI \\
Mini-SWE-agent & Mini-SWE-agent CLI \\
Terminus2 & Terminus2 terminal \\
\bottomrule
\end{tabular}
\caption{Overview of representative built-in agent harnesses available in
AgentCompass.}
\label{tab:harnesses_appendix}
\end{wraptable}

To interact with these diverse environments, the infrastructure provides a wide spectrum of agent harnesses, ranging from lightweight direct chat-completion wrappers to complex, in-environment autonomous frameworks (summarized in Table~\ref{tab:harnesses_appendix}). Complementarily, the built-in harnesses are designed to standardize the execution boundaries across fundamentally different agent architectures. This allows researchers to evaluate closed-source commercial tools (e.g., Claude Code, OpenAI Codex) alongside open-source autonomous frameworks (e.g., OpenHands, Mini-SWE-agent) under the exact same evaluation metrics and trajectory logging protocols. To facilitate specific research domains, the infrastructure also provides tailored workflows like the Naive Search Agent, a lightweight harness specifically developed for deep research that comes pre-equipped with search and web-browsing tools.

By leveraging a unified registry design, researchers can seamlessly pair any compatible benchmark with any harness, enabling standardized evaluation of heterogeneous agents without requiring additional glue code.
\section{Experiments}

\subsection{Experimental Settings}

To validate the efficacy, scalability, and universality of AgentCompass, we conduct a comprehensive evaluation of representative LLMs across eight highly challenging benchmarks. 
Specifically, the evaluated models include Qwen3.5-397B-A17B~\cite{qwen35blog}, DeepSeek-V4-pro(FP4)~\cite{xu2026deepseek}, Kimi-K2.6~\cite{moonshot2026kimik26}, GLM-5.2(FP8)~\cite{glm5team2026glm5vibecodingagentic}, GPT-5.5~\cite{openai2026gpt55}, Gemini-3.1-Pro-Preview~\cite{deepmind2026gemini31pro}, and Claude-Opus-4.8~\cite{anthropic2026claude48}. 
Model names followed by `()' indicate the corresponding quantized versions used in our evaluation.

The evaluation brings together a representative set of challenging benchmarks with their corresponding harnesses, including $\tau^3$-bench with its official workflow~\cite{shi2026tau}, DeepSearchQA and FrontierScience(Research)~\cite{gupta2026deepsearchqa,wang2026frontierscience} with Naive Search Agent, an in-house search harness developed in AgentCompass, SWE-bench-Pro and SWE-bench-Multilingual~\cite{deng2025swe,yang2025swesmith} with Mini-SWE-agent~\cite{yang2024sweagent} and OpenHands~\cite{wang2025openhands}, PinchBench~\citep{pinchbench2026skill} with OpenClaw~\cite{openclaw2026github}, SkillsBench~\cite{li2026skillsbench} with OpenClaw and OpenHands, and SciCode~\cite{tian2024scicode} with a multi-turn workflow equipped with official tools. 
All reported evaluation results are averaged over three independent runs to reduce measurement variance.
Further details are deferred to Appendix~\ref{sec:appendix_settings}.

\begin{table*}[t]
\centering
\definecolor{poscolor}{HTML}{006400}
\definecolor{negcolor}{HTML}{B22222}
\newcommand{\pos}[1]{\ensuremath{\,{}_{\color{poscolor}{\scriptstyle +#1}}}}
\newcommand{\nega}[1]{\ensuremath{\,{}_{\color{negcolor}{\scriptstyle -#1}}}}
\newcommand{\val}[2]{\makebox[4em][c]{#1\rlap{#2}}}
\newcommand{\hdr}[1]{\makebox[4em][c]{\fontsize{10}{12}\selectfont #1}}
\resizebox{\textwidth}{!}{
\renewcommand{\arraystretch}{1.2}
\begin{tabular}{l c c cc c cc cc cc}
\toprule
\multirow{3}{*}{\textbf{Model}} & \textbf{Tool Use} & \textbf{Web \& Research} & \multicolumn{2}{c}{\textbf{Scientific Reasoning}} & \multicolumn{3}{c}{\textbf{Productivity}} & \multicolumn{4}{c}{\textbf{Agentic Coding}} \\
\cmidrule(lr){2-2} \cmidrule(lr){3-3} \cmidrule(lr){4-5} \cmidrule(lr){6-8} \cmidrule(lr){9-12}
 & \multirow{2}{*}{$\tau^3$-bench} & \multirow{2}{*}{DeepSearchQA} & \multirow{2}{*}{FrontierSci} & \multirow{2}{*}{SciCode} & \multirow{2}{*}{PinchBench} & \multicolumn{2}{c}{SkillsBench} & \multicolumn{2}{c}{SWE-Pro} & \multicolumn{2}{c}{SWE-Multilingual} \\
\cmidrule(lr){7-8} \cmidrule(lr){9-10} \cmidrule(lr){11-12}
 & & & & & & \hdr{OpenClaw} & \hdr{OpenHands} & \hdr{MiniSWE} & \hdr{OpenHands} & \hdr{MiniSWE} & \hdr{OpenHands} \\
\midrule
Qwen3.5-397B-A17B      & 51.78 & 59.41 & 26.67 & 46.35 & 90.10 & \val{35.58}{} & \val{37.36}{} & \val{43.55}{} & \val{42.04}{} & \val{65.00}{\nega{4.3}} & \val{63.78}{\nega{5.5}} \\
Kimi-K2.6              & 50.76 & 71.52 & 48.27 & \val{51.87}{\nega{0.3}} & 87.35 & \val{53.10}{} & \val{50.62}{\nega{3.3}} & \val{58.09}{\nega{0.5}} & \val{61.29}{} & \val{77.78}{\pos{1.0}} & \val{77.22}{} \\
DeepSeek-V4-pro(FP4)   & 46.02 & 68.22 & 42.22 & 47.53 & 88.51 & \val{49.53}{} & \val{47.12}{\nega{2.9}} & \val{49.98}{\nega{5.4}} & \val{41.77}{\nega{13.6}} & \val{72.44}{\nega{3.8}} & \val{61.78}{\nega{14.4}} \\
GLM-5.2(FP8)           & 61.42 & 77.96 & 57.22 & 51.97 & 88.76 & \val{53.19}{} & \val{52.63}{} & \val{78.80}{} & \val{77.06}{\pos{15.0}} & \val{82.00}{} & \val{81.78}{} \\
Gemini-3.1-Pro-Preview & 48.22 & 53.00 & 25.00 & \val{54.44}{\nega{4.6}} & 87.97 & \val{44.99}{} & \val{44.64}{\nega{8.1}} & \val{39.26}{\nega{15.0}} & \val{45.69}{\nega{8.5}} & \val{44.00}{} & \val{63.00}{} \\
GPT-5.5                & 62.94 & 72.89 & 41.67 & 55.92 & 91.81 & \val{49.52}{} & \val{56.08}{\nega{11.2}} & \val{52.94}{\nega{5.6}} & \val{56.22}{\nega{2.4}} & \val{73.33}{} & \val{78.00}{} \\
Claude-Opus-4.8        & 55.33 & \val{76.11}{\nega{8.7}} & \val{36.67}{} & \val{56.21}{} & \val{90.70}{} & \val{54.40}{} & \val{58.66}{\pos{4.6}} & \val{66.21}{\nega{3.0}} & \val{73.87}{\pos{4.7}} & \val{77.00}{\nega{7.4}} & \val{77.00}{\nega{7.4}} \\
\bottomrule
\end{tabular}
}
\caption{Comprehensive evaluation results of representative models across the five core capability dimensions. Colored subscripts show gaps from official baselines, with green/red indicating higher/lower AgentCompass scores; no subscript means no official baseline was disclosed.}
\label{tab:main_results}
\end{table*}

\subsection{Main Results}

Table \ref{tab:main_results} shows that AgentCompass fairly aligns heterogeneous tasks and enables direct cross-harness comparisons, revealing that empirical agent capabilities are sensitive to the underlying harness. 
For instance, model performance fluctuates significantly depending on the employed harness, as evidenced by the discrepancies between OpenClaw and OpenHands on SkillsBench, or Mini-SWE-agent and OpenHands on the SWE-bench variants. 

Beyond absolute scores, the gap annotations in Table~\ref{tab:main_results} further show that agent evaluation is highly sensitive to infrastructure choices. 
Models can deviate substantially from their officially reported baselines under the unified AgentCompass protocol, such as Claude-Opus-4.8 dropping by 8.7 points on DeepSearchQA and GLM-5.2(FP8) improving by 15.0 points on SWE-bench-Pro with OpenHands. 
Such fluctuations may also arise from differences in harness versions, or from benchmark-specific adaptations made to the harness during evaluation. 
These discrepancies underscore the need for an open-source, unified evaluation framework that standardizes settings, implementations, and reporting protocols for reproducible and comparable agent assessment.

\subsection{Analysis Results}

In this section, we illustrate the importance of trajectory analysis in agentic task evaluation from several concrete perspectives.

\begin{wrapfigure}{r}{0.5\textwidth}
    \centering
    \includegraphics[width=\linewidth]{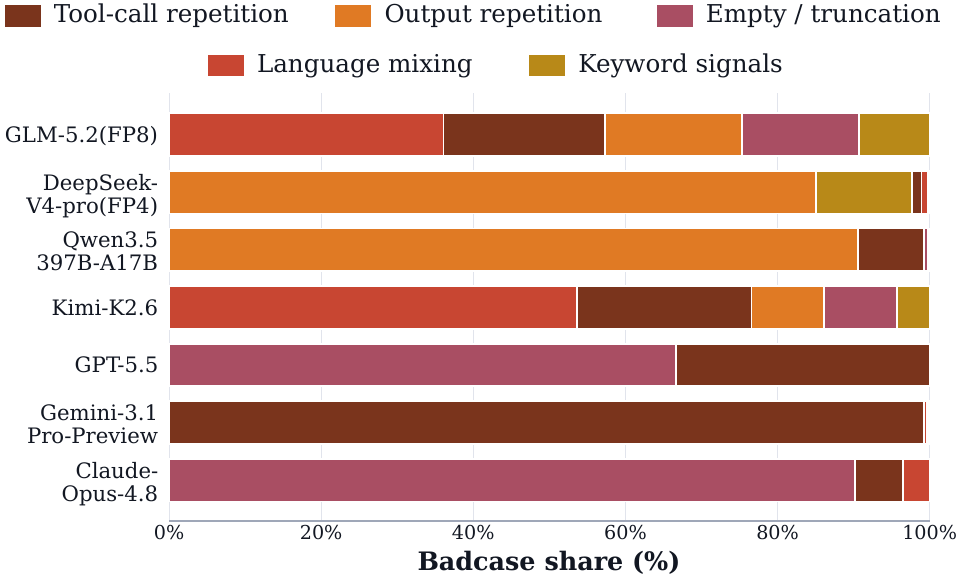}
    \caption{Distribution of bad-case behaviors across model trajectories.}
    \label{fig:badcase-distribution}
\end{wrapfigure}

\paragraph{RQ1: Beyond final scores, what types of behavioral bad cases occur in model trajectories?}
To characterize trajectory-level behavioral patterns, we leveraged the analysis capabilities of AgentCompass to quantify the distribution of bad-case behaviors. Figure \ref{fig:badcase-distribution} shows the distribution of behavioral bad-case patterns across models, where percentages indicate the proportion of each type among a model’s total bad cases. DeepSeek-V4-pro(FP4) mainly fail through repetitive content generation. Kimi-K2.6 has a higher share of bad cases involving multilingual mixing on search tasks and repeated tool calls. Gemini-3.1 is dominated by repeated tool calls, while Claude-Opus-4.8 and GPT-5.5 mainly produce empty outputs, although GPT-5.5 has very few bad cases overall. These findings suggest that, even when final task outcomes appear comparable, models may differ substantially in their underlying interaction dynamics and failure characteristics. Such beyond-score trajectory analysis therefore provides a more informative basis for diagnosing agentic behavior and identifying targeted directions for model improvement.

\begin{table*}[t]
\centering
\begin{subtable}[t]{0.49\textwidth}
\centering
\captionsetup{justification=centering}
\fontsize{8.5}{10}\selectfont
\begin{tabular}{lcc}
\toprule
Model & Sample-level & Step-level \\
\midrule
Qwen3.5-397B-A17B  & 3.08\%  & 0.06\%  \\
Kimi-K2.6 & 17.2\%& 0.73\%\\
DeepSeek-V4-pro(FP4) & 0.82\% & 0.06\% \\
GLM-5.2(FP8) & \textbf{39.12\%} & \textbf{2.09\%} \\
Gemini-3.1-Pro-Preview & 19\% & 1.05\% \\
GPT-5.5 & 17.03\%& 0.59\%\\
Claude-Opus-4.8 & 9.71\% & 0.65\% \\
\bottomrule
\end{tabular}%

\caption{SWE-Pro}
\label{tab:reward_hack_swepro}
\end{subtable}
\hfill
\begin{subtable}[t]{0.49\textwidth}
\centering
\captionsetup{justification=centering}
\fontsize{8.5}{10}\selectfont
\begin{tabular}{lcc}
\toprule
Model & Sample-level & Step-level \\
\midrule
Qwen3.5-397B-A17B  & 13.07\%  & 0.35\%  \\
Kimi-K2.6 & 20.51\%& 0.74\%\\
DeepSeek-V4-pro(FP4) & 4.02\% & 0.16\% \\
GLM-5.2(FP8) & 9.83\% & 0.24\% \\
Gemini-3.1-Pro-Preview & \textbf{21.97\%} & \textbf{0.80\%} \\
GPT-5.5 & 6.41\%& 0.37\%\\
Claude-Opus-4.8 & 5.63\% & 0.33\% \\
\bottomrule
\end{tabular}%

\caption{SWE-Multilingual}
\label{tab:reward_hack_swemulti}
\end{subtable}
\caption{Reward-hacking analysis results for different models using the Mini-SWE-agent harness.}
\label{tab:reward_hack_results}
\end{table*}

\paragraph{RQ2: Are high-scoring models on coding benchmarks genuinely strong?}
Using our reward-hacking analyzer inspired by \cite{deshpande2026benchmarkingrewardhackdetection}, we analyze correct samples on SWE-Pro and SWE-Multilingual with a two-level taxonomy of suspected hacking behaviors. Notably, the notion of suspected reward hacking adopted here is defined behaviorally rather than evidentially. Specifically, any action exhibiting characteristics of hacking is classified as reward hacking, regardless of whether there is direct evidence establishing a causal relationship between the behavior and the final outcome. Most models exhibit suspected reward hacking, such as modifying tests, retrieving golden patches. GLM-5.2(FP8) shows the highest suspected hacking rate on SWE-Pro, while Gemini-3.1-Pro-Preview does so on SWE-Multilingual; DeepSeek-V4-pro(FP4) remains consistently low on both. Notably, although GLM-5.2(FP8) outperforms Claude-Opus-4.8 by about 12 points on SWE-Pro, it also shows roughly 30\% more suspected reward-hacking samples. Detailed results are illustrated in Table \ref{tab:reward_hack_results}. These findings highlight the importance of carefully designing execution environments and constraints in both training and evaluation to ensure benchmark results better reflect coding ability.

\begin{figure*}[t]
    \centering
    \includegraphics[width=\textwidth]{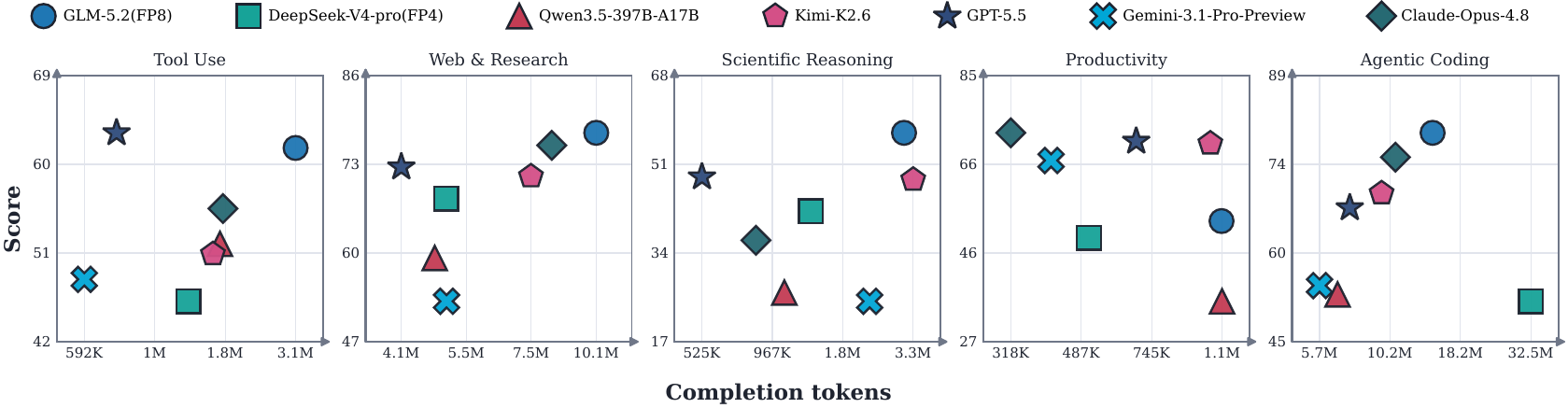}
    \caption{Capability--token length trade-offs across different capability dimensions.}
    \label{fig:capability_cost_all}
\end{figure*}

\paragraph{RQ3: How do model capability and token length relate across different tasks?} 
To help users make more scientifically grounded and operationally efficient decisions under a given budget, Figure \ref{fig:capability_cost_all}  illustrates the relationship between model performance and token length across different task categories. On coding tasks, most models follow the test-time scaling law, with longer outputs generally associated with higher scores, while DeepSeek-V4-Pro(FP4) appears as a notable outlier. On productivity tasks, Claude-Opus-4.8 stand out by achieving higher scores with lower token budgets, whereas Qwen3.5-397B-A17B performs relatively weaker. On tool-use and web-search tasks, GPT-5.5 tends to produce shorter outputs while maintaining comparatively strong performance.
Detailed statistics on the average interaction steps across benchmarks are provided in Appendix~\ref{sec:appendix_step}, which further characterize the execution complexity and interaction depth of different evaluation tasks.
\section{Conclusion}

In this paper, we introduced AgentCompass, an open-source and highly extensible infrastructure designed to systematize the evaluation of LLM-based agents. 
By decoupling the evaluation pipeline into independent Model, Benchmark, Harness, and Environment components, it eliminates redundant engineering and ensures rigorous reproducibility. 
Supported by a fault-tolerant asynchronous runtime and granular trajectory analysis, AgentCompass goes beyond traditional scalar scores to help researchers diagnose complex behaviors such as reward-hacking. 
With native integration of over 20 diverse benchmarks, AgentCompass provides a robust and scalable infrastructure to accelerate the next generation of agent research.

\section*{Acknowledgments}

This project is supported by Shanghai Artificial Intelligence Laboratory.

\section*{Author Contributions}
The authors are listed in alphabetical order by their last names: 

Kai Chen, Zichen Ding, Jiaye Ge, Shufan Jiang, Mo Li, Qingqiu Li, Zehao Li, Zonglin Li, Tianhao Liang, Shudong Liu, Zerun Ma, Zixin Shang, Wenhui Tian, Zun Wang, Liwei Wu, Zhenyu Wu, Jun Xu, Bowen Yang, Dingbo Yuan, Qi Zhang, Songyang Zhang, Peiheng Zhou, Dongsheng Zhu

\clearpage
\bibliographystyle{plain}
\bibliography{refs}


\clearpage
\appendix
\section{Detailed Experimental Configurations}
\label{sec:appendix_settings}

In this section, we provide the detailed experimental settings used in our evaluation to ensure full reproducibility. 

\subsection{Model Configurations} 
For closed-source models (e.g., GPT-5.5, Claude-Opus-4.8), we used the default inference settings provided by the corresponding model APIs to ensure deterministic outputs where possible. 
For open-weight models (e.g., Qwen3.5-397B-A17B), we deployed them with the SGLang inference engine and followed the official recommended inference configurations for each model. 
For all evaluated models, whenever a reasoning-effort parameter was available, we set it to \texttt{high}. 
This consistent setup avoids ad hoc parameter tuning and keeps the reported comparisons reproducible.

\subsection{Benchmark Specification.}
In Table~\ref{tab:main_results}, the subscripted differences report the gap between the AgentCompass score and the closest available external reference result for the same model and benchmark.
External references are drawn from the model's official technical report and the benchmark's official leaderboard when available.
If both sources are available, we use the one whose reported score is closest to the AgentCompass evaluation result.
If neither source reports a comparable result, no subscripted difference is shown.

\paragraph{Tool Use, Web \& Research and Scientific Reasoning.}
For tool-use evaluation, we followed the official workflow of $\tau^3$-bench. 
In particular, both the user model and the rank model were instantiated with the model under evaluation, so that the evaluated model acted as both the conversational user simulator and the knowledge re-ranker. 
This configuration provides a more holistic evaluation of the model's end-to-end capabilities in interactive tool-use settings. 
For web \& research and scientific-reasoning evaluation, DeepSearchQA and FrontierScience(Research) were evaluated with the AgentCompass Naive Search Agent, whereas SciCode was evaluated with the AgentCompass port of the official tool-assisted multi-turn workflow. 
DeepSearchQA used Qwen3.6-35B-A3B as the judge model, while FrontierScience(Research) used GPT-5.5. 
For SciCode, we used the official reference checkout e3158e and loaded the task package hosted by AgentCompass. 
Scientist-annotated background information was enabled during evaluation. 
We evaluated all 80 problems, which resulted in 338 scored subproblems after excluding the three official prefilled steps. 
The SciCode harness ran in \texttt{tool\_use} mode with \texttt{code\_interpreter} as the only enabled tool, allowing up to 30 tool-use iterations per generated step and imposing a 180-second timeout for each local code-interpreter execution.

\paragraph{Agentic Coding.}
For software-engineering evaluation, SWE-bench Pro and SWE-bench Multilingual were evaluated with both Mini-SWE-agent (v2.3.0) and OpenHands, where OpenHands was instantiated through the OpenHands Software Agent SDK (v1.23.0). 
Both agents operated on the same repository checkout and were scored through an identical patch-generation interface and unit-test-based metric. 
The system prompt was supplied by the corresponding harness, and the user prompt consisted of the benchmark issue description together with an instruction to write the generated patch to \texttt{patch.txt}. 
For SWE-bench Pro instances based on minimal container images, such as Alpine Linux or Go toolchain images, we applied runtime compatibility fixes at container startup, including the provisioning of glibc-related components, and executed OpenHands inside a micromamba-managed isolated environment.

\paragraph{Productivity.}
For productivity-oriented evaluation, we used the official benchmark releases whenever available and kept task files and judging rules fixed across harnesses. 
SkillsBench used the AgentCompass data version derived from official commit 17dec32 and was evaluated with the integrated OpenClaw (v2026.5.9) and OpenHands (v1.23.0) harnesses under a task-level timeout multiplier of 24. 
PinchBench used the official skill repository release v1.1.0, corresponding to commit e8e833, and was evaluated on the full \texttt{suite=all} setting with 23 tasks using the AgentCompass-integrated OpenClaw harness. 
Unless otherwise specified, PinchBench runs used OpenClaw (v2026.3.22) with a 250000-token context window, an 80000-token per-turn limit, and a maximum single-message length of 131072 characters. 
For Kimi-K2.6, we used OpenClaw (v2026.7.1-beta.2) because the default version did not robustly handle responses with non-empty reasoning content but empty assistant content. 
All PinchBench runs were judged by \texttt{claude-opus-4-5-20251101} with thinking mode enabled; task timeouts followed the benchmark-defined \texttt{timeout\_seconds} values with a 10x multiplier, and the grading runner used a 480-second outer timeout to accommodate judge execution and result parsing.

\section{Step statistics}
\label{sec:appendix_step}

\begin{table*}[t]
\centering
\definecolor{poscolor}{HTML}{006400}
\definecolor{negcolor}{HTML}{B22222}
\newcommand{\pos}[1]{\ensuremath{\,{}_{\color{poscolor}{\scriptstyle +#1}}}}
\newcommand{\nega}[1]{\ensuremath{\,{}_{\color{negcolor}{\scriptstyle -#1}}}}
\newcommand{\val}[2]{\makebox[4em][c]{#1\rlap{#2}}}
\resizebox{\textwidth}{!}{    
\renewcommand{\arraystretch}{1.2}
\begin{tabular}{l c c c c c c c c}
\toprule
\textbf{Model} & $\tau^3$-bench & DeepSearchQA & FrontierSci & SciCode & PinchBench & SkillsBench & SWE-Pro & SWE-Multilingual \\
\midrule
Qwen3.5-397B-A17B      & 21.56 & 19.21 & 4.94 & 4.26 & 5.26 & 34.45 & 59.42 & 68.31 \\
Kimi-K2.6              & 20.68 & 23.36 & 8.73 & 4.29 & 5.29 & 38.50 & 57.72 & 50.85 \\
DeepSeek-V4-pro(FP4)   & 18.40 & 15.37 & 3.49 & 4.06 & 4.26 & 30.83 & 44.35 & 38.55 \\
GLM-5.2(FP8)           & 18.06 & 16.23 & 9.41 & 4.13 & 5.10 & 41.76 & 67.76 & 60.59 \\
Gemini-3.1-Pro-Preview & 48.14 & 27.44 & 9.01 & 4.00 & 4.77 & 43.10 & 76.29 & 99.56 \\
GPT-5.5                & 20.07 & 21.60 & 11.16 & 4.26 & 6.57 & 33.18 & 46.59 & 31.64 \\
Claude-Opus-4.8        & 17.92 & 10.03 & 1.74 & 4.16 & 5.13 & 21.16 & 30.22 & 19.13 \\
\bottomrule
\end{tabular}
}
\caption{Average interaction steps per evaluation instance across different benchmarks. Each value represents the mean trajectory length over all evaluated samples for the corresponding model and benchmark.}
\label{tab:step_statis}
\end{table*}

Table~\ref{tab:step_statis} summarizes the average number of interaction steps required to complete one evaluation instance across different benchmarks.
Here, a \emph{step} denotes one complete agent--environment interaction cycle, including one model generation together with the corresponding tool invocation (if any) and the returned environment observation.
For purely conversational benchmarks, a step is equivalent to one model response, whereas for interactive coding or productivity benchmarks, a step typically consists of one reasoning action followed by environment execution.

The statistics are computed by averaging the recorded trajectory lengths over all evaluated instances for each model and benchmark.
Because AgentCompass records the complete execution trajectory in a unified format, these values are collected automatically without benchmark-specific instrumentation.
Several observations can be drawn from Table~\ref{tab:step_statis}.
First, benchmarks exhibit substantially different interaction depths.
Scientific reasoning tasks (e.g., \textsc{SciCode}) generally require only around four interaction steps due to their constrained tool workflow, whereas software engineering benchmarks frequently require tens of iterations because agents repeatedly inspect repositories, modify source files, execute tests, and refine patches.
Second, model-specific differences are also evident.
For example, Gemini-3.1-Pro-Preview performs considerably longer trajectories on the SWE benchmarks than other models, while Claude-Opus-4.8 typically completes tasks with fewer interaction steps.
Finally, longer trajectories do not necessarily translate into higher benchmark scores.
As illustrated in the main paper (Figure~4), the relationship between interaction length and evaluation performance is benchmark dependent.
These trajectory-level statistics therefore provide complementary evidence beyond final task accuracy and are useful for understanding evaluation cost, interaction efficiency, and agent behavior.



\end{document}